\documentclass[sigconf]{acmart}

\usepackage{booktabs}
\usepackage{xspace}
\usepackage{graphicx}
\usepackage{caption}
\usepackage{subcaption}
\usepackage{tikz}

\renewcommand\footnotetextcopyrightpermission[1]{} 
\usetikzlibrary{positioning}
\usetikzlibrary{decorations.pathreplacing}
\usetikzlibrary{arrows}
\usetikzlibrary{decorations,arrows}
\usetikzlibrary{decorations.pathmorphing}
\usetikzlibrary{fit}
\usetikzlibrary{patterns}

\setcopyright{rightsretained}

\begin{document}
\title{Federated Learning for Ranking Browser History Suggestions}

\acmDOI{?}
\acmISBN{?}
\acmArticle{?}
\acmPrice{?}

\acmConference{}
\acmYear{2019}
\copyrightyear{2019}

\author{Florian Hartmann}
\authornote{Now at Google Research.}
\affiliation{\institution{Mozilla}}
\email{fhartmann@google.com}

\author{Sunah Suh}
\affiliation{\institution{Mozilla}}
\email{ssuh@mozilla.com}

\author{Arkadiusz Komarzewski}
\affiliation{\institution{Mozilla}}
\email{akomarzewski@mozilla.com}

\author{Tim D. Smith}
\affiliation{\institution{Mozilla}}
\email{tdsmith@mozilla.com}

\author{Ilana Segall}
\affiliation{\institution{Mozilla}}
\email{isegall@mozilla.com}

\renewcommand{\shortauthors}{F. Hartmann et al.}

\newcommand{\fl}{Federated Learning\xspace}

\begin{abstract}

Federated Learning is a new subfield of machine learning that allows fitting models without collecting the training data itself.
Instead of sharing data, users collaboratively train a model by only sending weight updates to a server.
To improve the ranking of suggestions in the Firefox URL bar, we make use of Federated Learning to train a model on user interactions in a privacy-preserving way.
This trained model replaces a handcrafted heuristic, and our results show that users now type over half a character less to find what they are looking for.

To be able to deploy our system to real users without degrading their experience during training, we design the optimization process to be robust.
To this end, we use a variant of Rprop for optimization, and implement additional safeguards.
By using a numerical gradient approximation technique, our system is able to optimize anything in Firefox that is currently based on handcrafted heuristics.
Our paper shows that \fl can be used successfully to train models in privacy-respecting ways.
\end{abstract}

%
%

\ccsdesc[500]{Computing methodologies~Machine learning}
\ccsdesc[300]{Computing methodologies~Distributed artificial intelligence}
\ccsdesc[100]{Computing methodologies~Learning to rank}
\ccsdesc[100]{Computing methodologies~Distributed algorithms}

\keywords{federated learning, privacy-preserving machine learning, learning to rank}

\maketitle

\section{Introduction}
Mozilla is constantly balancing user privacy with the benefits that could be produced for users by collecting data.
There are processes in place to ensure that all data collected is reviewed and evaluated against principles of necessity, privacy, transparency, and accountability~\cite{firefox-data}.
When it comes to developing machine-learning-based products, however, many potential ideas would traditionally have required collecting highly sensitive user data.
Even if some users agree to share their data, it is difficult to build representative models in this way.

A promising solution to Mozilla's problem is \emph{\fl}, a new technique that allows training models on data which only exists on user devices.
Rather than sharing their data, users only send model improvements to a server.
These updates are locally derived based on private data.
Since the data is not directly shared, and there exist various additional protection mechanisms in \fl, this approach offers much better privacy.

To experiment with \fl, we decided to use it to try to improve the Firefox URL bar.
When typing a query into the URL bar, Firefox tries to suggest history entries, bookmarks, and other items that the user might want to click on.
At the time of our experiment, the algorithm that selects and ranks these suggestions was based on handcrafted heuristics.
It used several constants for configuring how different features should be weighted.
The values of these weights were chosen because they intuitively seemed reasonable, and not because a data-driven process provided evidence that they work well.

Since many users regularly type queries into the URL bar and select from the suggestions shown to them, Firefox could collect data in order to optimize the URL bar with a learning algorithm.
However, search queries, history entries, and bookmarks are clearly private to users.
Collecting this data on a server would severely violate privacy constraints or produce non-representative results due to opt-in bias.

To still optimize the URL bar suggestions based on user interactions, we developed a system that uses \fl.
This system allows us to compute model improvements based on the users' data, without directly collecting it.
Instead, the optimization algorithm is distributed so that the parts that directly touch the user data are also executed on the users' computers.

To the best of our knowledge this system represents the first use of \fl in a major software product outside of Google.
\fl research published so far has been based purely on simulations.
These simulations represent an easier problem setting since the developer can test many different optimization algorithms and hyperparameters in a short amount of time.

Deploying a \fl system is more difficult because there is a much larger cost attached to testing out different versions.
Each experiment with real users takes much more time, and, even worse, can lead to a bad user experience.
For these reasons, we developed additional techniques to make our system robust enough to get good results by deploying it to Firefox just once.
These contributions include carefully designing the optimization algorithm, and implementing safeguards to ensure model quality.

We also designed our system in a way that it can optimize anything in Firefox that is currently based on hardcoded constants.
Since there are other places where handcrafted heuristics could be replaced by trained models, we believe our system to be widely applicable in Firefox.
All code that was used during the experiment, and in preparing it, is open sourced in accordance with Mozilla's philosophy~\cite{impl-client, impl-server, impl-simulations}.

Our system ended up being deployed to a large fraction of Firefox Beta users.
Over the course of under four days, roughly 360,000 users helped to train and evaluate a new model for the improved URL bar.
The new model leads to users typing over half a character less before selecting one of the proposed suggestions.
We have released an anonymized dataset containing the client data collected during this study [forthcoming].

\begin{figure}[h]
    \centering
    \includegraphics[width=250px]{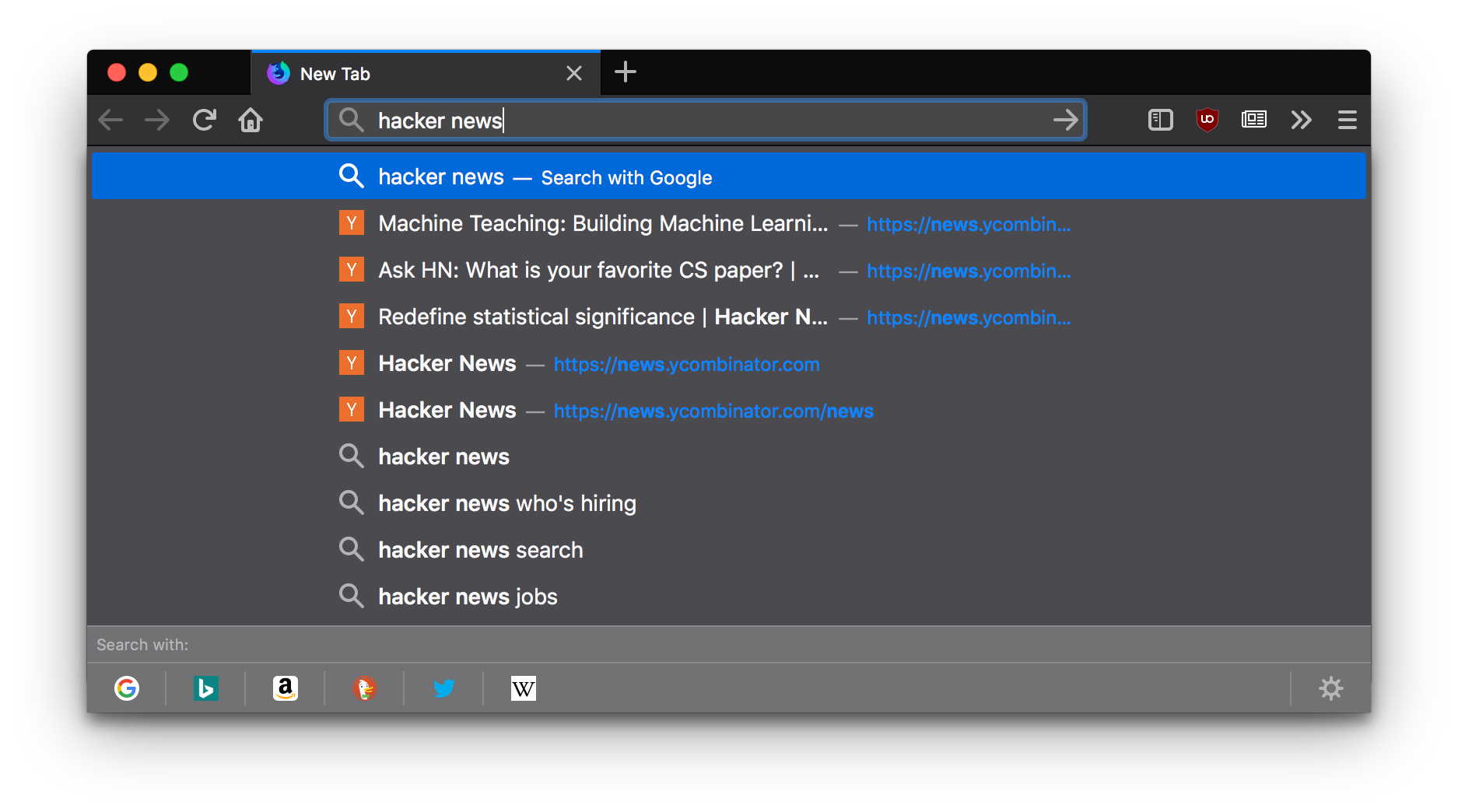}
    \caption{The Firefox URL bar provides suggestions based on the browsing history}
    \label{fig:fx-bar}
\end{figure}

In this paper, we describe how we built the system, what we learned from doing so, and how it could be improved in the future.
Concretely, our contributions include:
\begin{itemize}
    \item We show that \fl can be used successfully in major software products, rather than just in simulations.
    \item Since little control over data can be exercised in this setting, we propose specialized optimization methods.
    \item We discuss how any black box functions can be optimized based on user interactions, without sacrificing user privacy, by using a custom way of computing gradients.
\end{itemize}

The remainder of the paper is structured as follows.
Section~\ref{sec:related-work} briefly introduces related work, while Section~\ref{sec:background} gives the necessary background knowledge.
In Section~\ref{sec:fx-search}, we show how the URL bar search in Firefox generally works.
Section~\ref{sec:optimization} explains how we designed the optimization process to be robust and easily reusable for future applications.
The process of launching our experiment as well as the analysis of the results are discussed in Section~\ref{sec:study}.
Finally, we provide a short conclusion in Section~\ref{sec:conclusion}.

\section{Related Work}
\label{sec:related-work}

\citet{fl} initially proposed \fl and an algorithm for computing model updates in a distributed way.
This algorithm repeatedly receives gradients from clients, averages them and applies them to the model.

Since neural networks can contain millions of parameters, this approach could lead to a high communication cost.
To make the algorithm feasible for large models, various compression techniques have been suggested subsequently~\cite{compression}.

\section{Background}
\label{sec:background}

\subsection{Federated Learning}
\label{subsec:fl}

One of the reasons why machine learning has been applied so successfully to many problems in the past few years is that more data has become available.
For some applications of machine learning, collecting data can be privacy-invasive.
One such example application is predicting the next word that a person is going to use by considering the previously typed words.
This is typically done using machine learning nowadays, e.g.\ with recurrent neural networks and LSTMs~\cite{lstm}.

Although it is possible to train such a model using a text corpus from Wikipedia, the language found there differs from the one commonly used by people in daily life.
To train a model on the same data distribution that is also used for inference when the model is deployed, one would need to collect data directly from users.
This, however, would violate the privacy of users.
Users do not want to send everything they type to a server.

Federated Learning is a recently proposed technique for training models on this data, without sending it to a server.
Instead, we collaboratively train a model by distributing the training process among many users.
A server has the role of coordinating this process but most of the work is not performed by a central entity anymore but by a \emph{federation} of users.

Before the server starts off the distributed learning process, it needs to initialize the model.
Theoretically, this can be done arbitrarily, by using any of the common model initialization strategies.
In practice, it makes sense to intelligently initialize the model with sensible default values.
If some data is already available on the server, it can be used to pretrain the model.
In other cases, there might be a known configuration of model parameters that already leads to acceptable results.
Having a good first model gives the training process a head start and can reduce the time until convergence is reached.

\begin{figure*}[ht]
\begin{subfigure}{.45\textwidth}
 \centering
 \includegraphics[width=.8\linewidth]{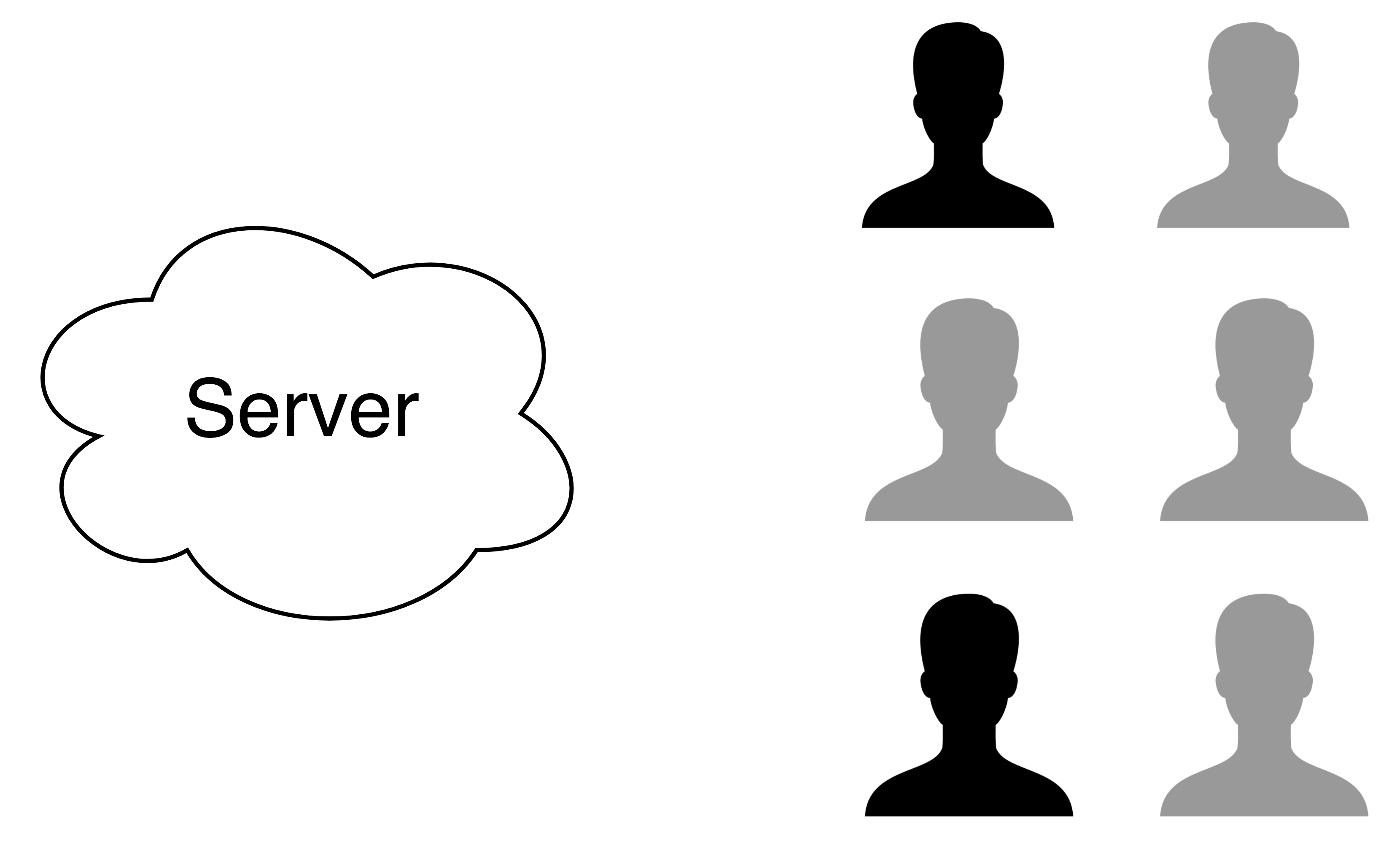}
 \caption{The server selects $K$ users}
 \label{fig:sfig2}
\end{subfigure}
\begin{subfigure}{.45\textwidth}
 \centering
 \includegraphics[width=.8\linewidth]{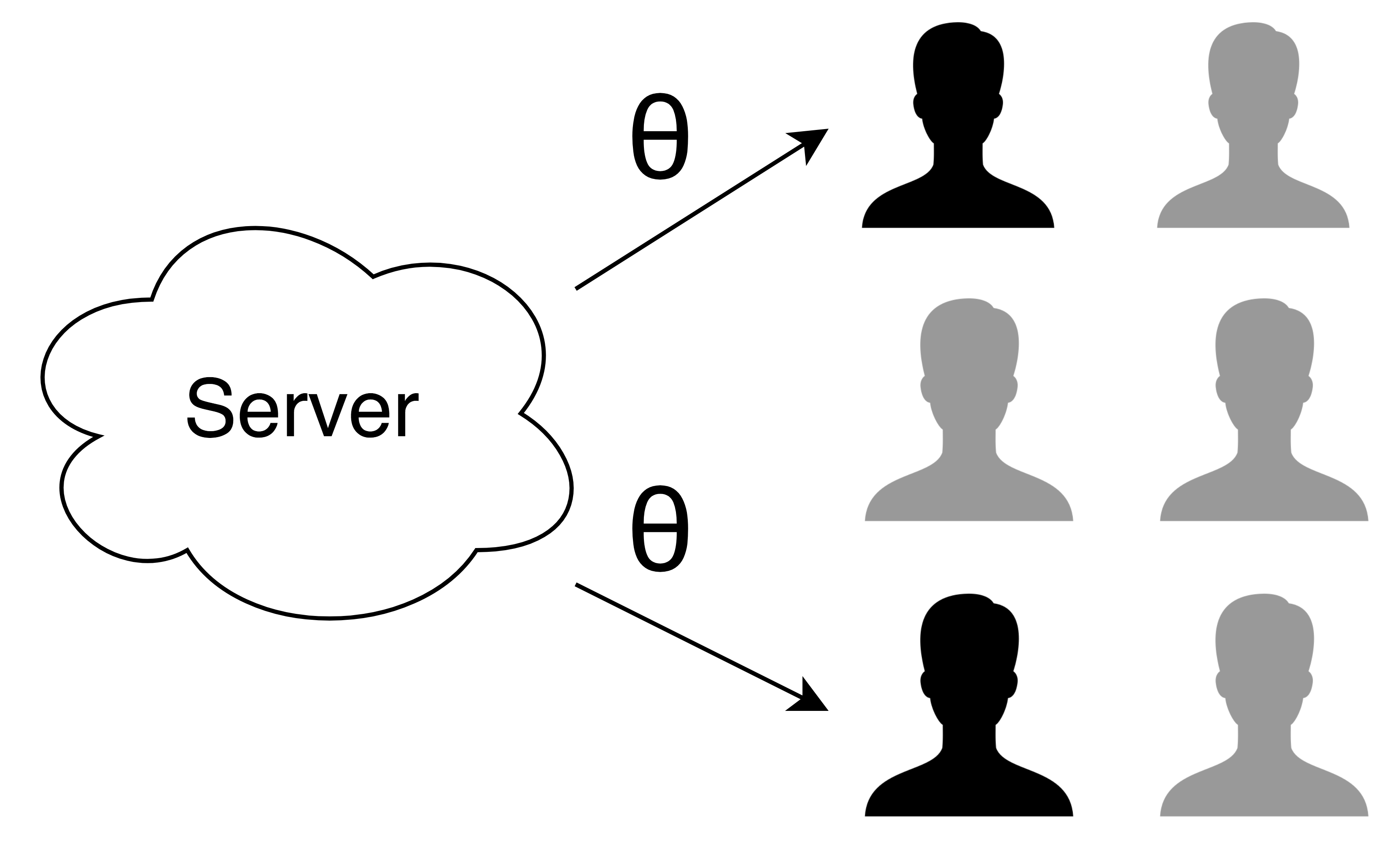}
 \caption{They receive the current model}
 \label{fig:sfig2}
\end{subfigure}

\vspace{1cm}

\begin{subfigure}{.45\textwidth}
 \centering
 \includegraphics[width=.8\linewidth]{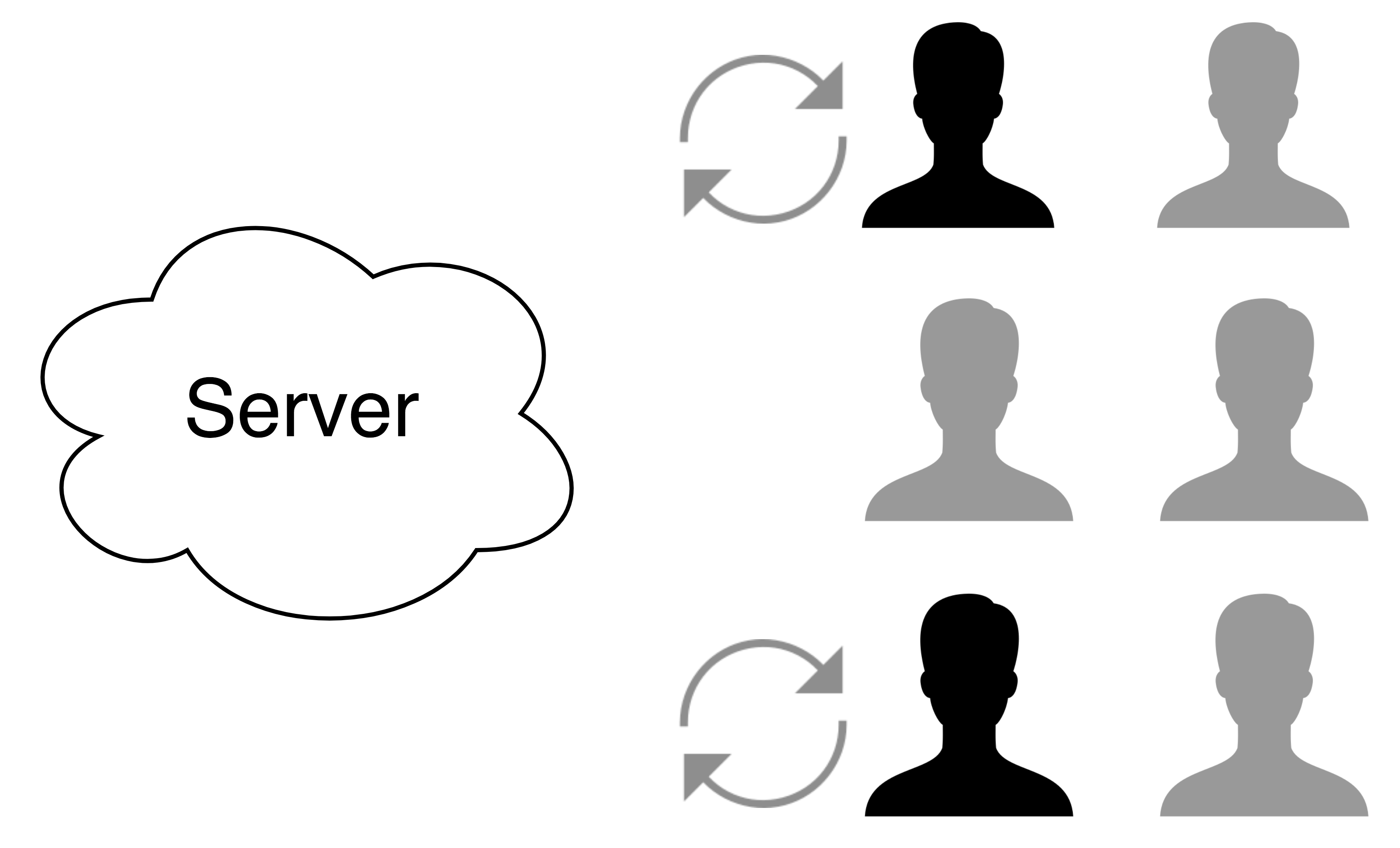}
 \caption{and compute updates using their data}
 \label{fig:sfig2}
\end{subfigure}
\begin{subfigure}{.45\textwidth}
 \centering
 \includegraphics[width=.8\linewidth]{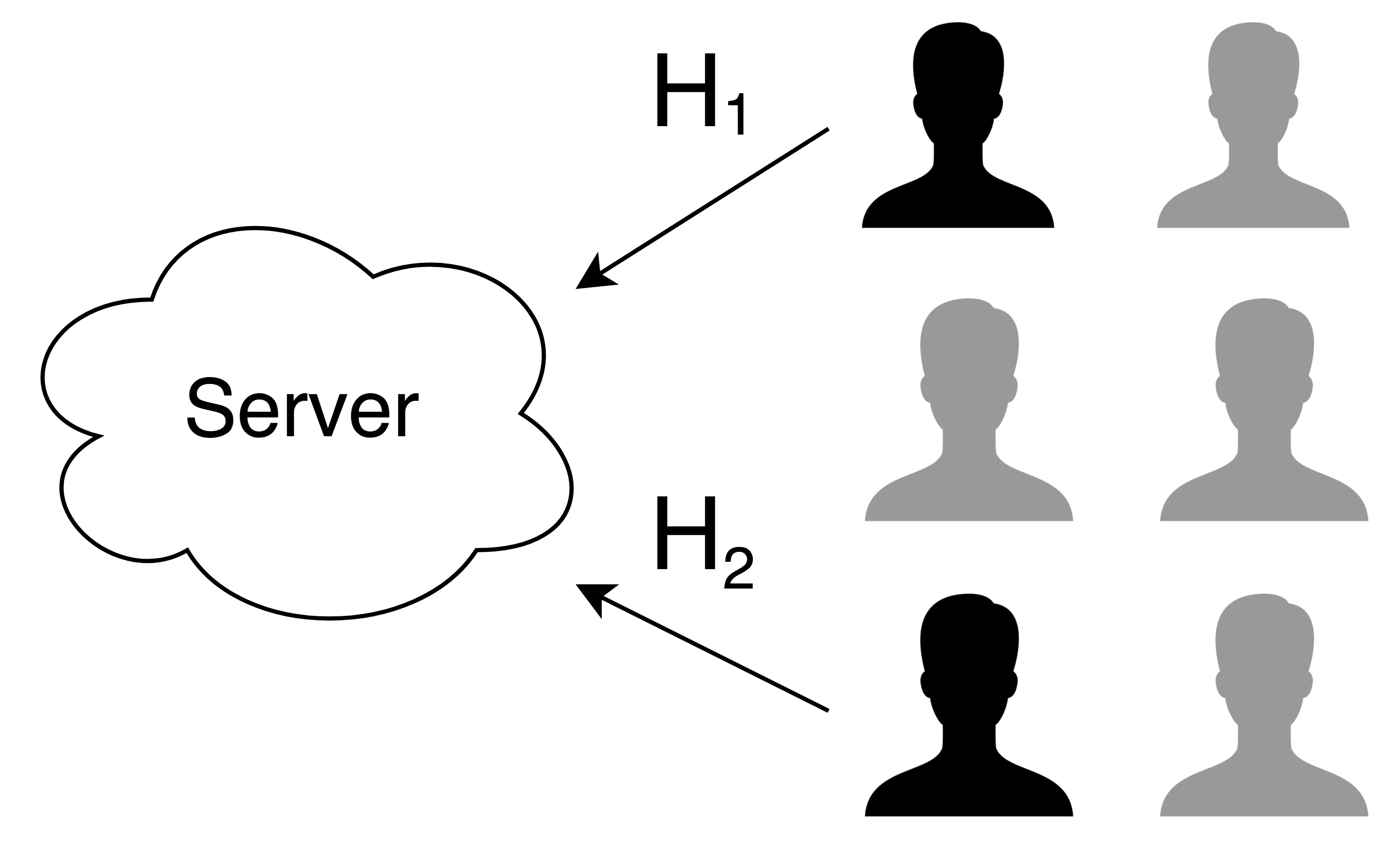}
 \caption{Updates are shared with the server}
 \label{fig:sfig2}
\end{subfigure}

\vspace{0.2cm}

\caption{One communication round in a \fl system}
\label{fig:fl}
\end{figure*}

After the model has been initialized, the iterative training process is kicked off.
At the beginning of an iteration, a subset of $K$ clients is randomly selected by the server.
They receive a copy of the current model parameters and use their locally available training data to compute an update.
The updates are then sent back to the server.
We denote the model parameter tensor by~$\theta$, the update of the \mbox{$i$-th} user by $H_i$, and the number of data points on the computer of the $i$-th user by $n_i$.
A visualization of the steps performed in each iteration is shown in Figure~\ref{fig:fl}.

The server waits until it has received all updates of the iteration and then combines them into one final update.
This is usually done by computing an average of all updates, weighted by how many training examples the respective clients used.
The weights for iteration $t$ are then computed using
\begin{equation*}
	\theta^{(t)} = \theta^{(t - 1)} - \sum\limits_{i = 1}^K \frac{n_i}{N} H_i
	\label{eq:federated-updates}
\end{equation*}
\noindent where $N = \sum_{i = 1}^K n_i$ is the total number of data points used in this round.
A new iteration begins after every model update.

In each iteration only $K$ users are queried for updates.
While requesting updates from all users would lead to more stable model improvements, it would also be extremely expensive to do because there could be millions of users.
Only querying a subset of them makes it more feasible to efficiently run many iterations.

This training process is then repeated until the model parameters converge, as determined by an appropriate criterion.
In some situations, it can also make sense to keep the training running indefinitely.
This can for example be the case in some recommender system applications, where the model needs to deal with new data all the time.

One potential attack vector against \fl systems is trying to analyze updates sent to the server, in order to make conclusions about the original data.
To prevent such \mbox{man-in-the-middle} attacks, \emph{secure multi-party aggregation} mechanisms can be used~\cite{encryp, encryp2}.
These protocols use encryption techniques to allow only the analysis of the aggregated updates from a certain number of users, thus securing individual updates.

To ensure that the model itself does not implicitly memorize personal information, \emph{differential privacy}~\cite{dp-foundations} offers a framework to formalize privacy.
Differential privacy has been applied to \fl, and there exist gradient descent variants that adhere to a specified level of $(\epsilon, \delta)$-differential privacy~\cite{noisy-sgd,fl-language-model}.

\subsection{Learning to Rank}
\label{subsec:ranking}

A ranking algorithm takes a set of items and sorts them by some criterion.
Before diving into ranking in the Firefox URL bar, it is worth taking a step back to understand how ranking in machine learning works.
This makes it easier to see how the current implementation fits into a learning system.
Fundamentally, there are three different approaches to learning a ranking algorithm~\cite{ranking1,ranking2}.

\emph{Pointwise ranking} algorithms process all items individually and assign a score to each item independently of the others.
The ranking is then determined by sorting all items using their respective scores.
Essentially, this is a special type of a regression model since we are assigning a real number to each input.

A \emph{pairwise ranking} model learns to compare pairs of items.
Its task is to decide which of the two items should be ranked higher.
The learned comparison function can then be used to sort the items.
In this approach, we treat the problem as a classification task since the model can only have two possible outputs.

The third approach is \emph{listwise ranking}, which are methods that try to operate on the entire list.
The motivation behind this idea is to optimize information retrieval metrics directly.
In practice, this turns out to be fairly difficult because many of those metrics are not differentiable and the models need to work with more inputs.

All these approaches have different advantages and disadvantages.
The existing ranking algorithm in Firefox is very similar to a pointwise ranking approach.
Since this algorithm should be optimized using machine learning techniques, this gives us a clear set of techniques that could be useful in this project.
To optimize the current algorithm, we took a known pairwise ranking loss function and adapted it to work with our pointwise ranking system.

\section{Search in the Firefox URL bar}
\label{sec:fx-search}

The Firefox URL bar offers suggestions when users type a search query.
A part of these suggestions is provided directly by a search engine.
The others are generated by Firefox itself, for example based on the user's history, bookmarks, or open tabs.
We tried to optimize the history and bookmark suggestions using our project.
The search engine results are shown below those, and we have no influence over their selection and ranking since everything is provided directly by the search engine.

Searching for history and bookmark entries in the Firefox URL bar is a two-step process:
\begin{enumerate}
    \item The search query is matched against the browser history and bookmarks. Matching is a binary decision. Pages either match the query or do not.
    \item The set of matched links is ranked based on the user's history.
\end{enumerate}

Our project purely tries to optimize the ranking part of this process.
Future work could tackle the problem directly from the query matching.

The ranking of possible suggestions in the Firefox URL bar is determined using \emph{frecency}~\cite{frecency}, an algorithm that weights how \emph{frequently} and \emph{recently} a site was visited.
To do this, a frecency score is assigned to each history entry and bookmark entry.
After the score is computed, it is cached.
When searching, the matched results are then sorted using their frecency scores.
This makes frecency a pointwise ranking approach.

Frecency does not only take frequency and recency into account but also other information, such as how the page was visited and whether it is bookmarked.
It does this by looking at the latest visits to the respective site.
The value $\operatorname{visit}(v)$ of one single visit $v$ is then defined by how recent that visit was, scaled by the type of visit:
$$
\operatorname{visit}(v) = \operatorname{recency}(v) * \operatorname{type}(v)
$$

Frecency scores have to be cached in order to allow an efficient ranking while the user is typing.
This means that the recency aspect has to be modeled using time buckets.
Otherwise, the score would change all the time and caching would not work.
In the current Firefox implementation, there are five time buckets.
With this approach, the recency score only changes when a visit changes time buckets:
$$
\operatorname{recency}(v) = \begin{cases}
	100 & \text{if visited in the past 4 days} \\
	 70 & \text{if visited in the past 14 days} \\
	 50 & \text{if visited in the past 31 days} \\
	 30 & \text{if visited in the past 90 days} \\
	 10 & \text{otherwise}
\end{cases}
$$

Sites can be visited in many different ways.
If the user typed the entire link themselves or if it was a bookmarked link, we want to weight that differently to visiting a page by clicking a link.
Other visit types, like some types of redirects, should not be worth any score at all.
We implement this by scaling the recency score with a type weight:
$$
\operatorname{type}(v) = \begin{cases}
	1.2 & \text{if URL was visited by following a link on a website} \\
	2   & \text{if URL was typed out} \\
	1.4 & \text{if URL is bookmarked} \\
	0   & \text{otherwise}
\end{cases}
$$

Now that we can assign a score to every visit, we could determine the full points of a page by summing up the scores of each visit to that page.
This approach has several disadvantages.
For one, it would scale badly because some pages are visited a lot.
Additionally, user preferences change over time and we might want to decrease the points in some situations.

Instead, we compute the average score of the last 10 visits.
This score is then scaled by the total number of visits.
The full frecency score can now be computed efficiently and changes in user behavior are reflected fairly quickly.
Let $S_x$ be the set of all visits to page $x$, and let $T_x$ be the set of the last up to 10 of these.
The full frecency score is then given by:
$$
\operatorname{frecency}(x) = \frac{|S_x|}{|T_x|} * \sum\limits_{v \in T_x} \operatorname{visit}(v)
$$

For legacy reasons, Firefox additionally decays frecency scores over time.
Once a day, all scores are decreased by a few percentage.
This feature exists in Firefox for historic reasons and could also be modeled using time buckets.

Note that this is a simplified version of the algorithm.
There is some additional logic for special cases, such as typing out bookmarks or different kinds of redirects.
The description here only shows the essence of the algorithm in a mathematical form.

The weights of the current algorithm were handchosen in a way they were deemed reasonable.
Our \fl system tries to optimize the weights based on user interactions, while keeping the general logic of frecency intact.
This optimization process includes all constants in the previous formulas.

\section{Optimization System Design}
\label{sec:optimization}

\subsection{Optimization By User Interactions}

User interactions with the Firefox URL bar provide a feedback signal that can be used to optimize the weights in the frecency algorithm.
By checking what items users clicked on, the weights can be adapted to make it more likely to show these items earlier the next time similar searches are performed.

The general optimization loop works as follows:
Users search in the URL bar and click on a suggestion.
This provides data to our system that can be used to compute model improvements, in the form of gradients.
These gradients are computed for many users at the same time and are sent to a Mozilla server.
A job on the server averages all gradients of the current iteration and applies a scaled version of the result to the model.
The new model is redistributed periodically when a new iteration begins.

Part of the promise of our system is that this feedback signal from users is perfectly clean.
Users do not only generate data by searching in the URL bar, but they also label it for ranking by selecting an item.
Thus we know exactly how the ranking should have been.
Since machine learning is relying on good data to work with, this is an important point.

\subsection{Pointwise SVM Ranking}

To describe the ranking goal formally, we need to have a loss function that evaluates how well our model did.
To this end, we take a previously proposed SVM loss for pairwise ranking~\cite{svm-ranking,svm} and adapt it to our pointwise approach.

Essentially, the ranking loss function takes in a set of items with their assigned scores as well as the index of the item selected by the user.
The optimization goal is that the selected item should have the highest score.

But even if this was the case, our model might not have been too confident in that decision.
One example of this is the selected item having a score of 100 and the second item having a score of 99.9.
The model made the correct prediction, but only barely so.
To make sure it does a good job in similar cases, we need to provide a signal to the model which shows that it can still improve.
This is what we aim to do with the SVM loss.

If the URL bar displayed the suggestions for pages $x_1, \dots, x_n$ in that order and suggestion $x_i$ was chosen, then the SVM loss for the pointwise ranking is given by
$$
E = \sum\limits_{j \neq i} \max(0, f(x_j) + \Delta - f(x_i))
$$
\noindent where $f(x_i)$ denotes the pointwise ranking score of item $x_i$, which corresponds to frecency in our case.

We iterate over all suggestions that were not chosen and check that their scores were smaller than the one of the selected page by at least a margin of $\Delta$.
If not, an error is added.
The full loss should be minimized.
The margin $\Delta$ is a hyperparameter that needs to be decided on before the optimization process starts.

\begin{figure}[H]
  \begin{subfigure}{.5\textwidth}
	\centering
	\begin{tikzpicture}[baseline=0]
		\fill[black] (-4, 0) rectangle (-3.5, 0.5);
		\node at (-2.5, .25) {selected item};
		\draw[pattern=north west lines, pattern color=black] (-1, 0) rectangle (-0.5, 0.5);
		\node at (-0.05, .25) {loss};
	\end{tikzpicture} \\
\end{subfigure}

  \begin{subfigure}{0.5\textwidth}
    \centering
	\begin{tikzpicture}[samples=100,smooth,scale=0.75]
		\begin{scope}
        	\draw[->,ultra thick] (-5,-4.2)--(5,-4.2) node[right]{$x_i$};
			\draw[->,ultra thick] (-4.7,-4.5)--(-4.7,3.0) node[above]{$f(x_i)$};

			\draw[black] (-4.7,-4.2) rectangle (-3.7,1);
			\draw[pattern=north west lines, pattern color=black] (-4.7, 0) rectangle (-3.7,1);
			\draw[black] (-3.7,-4.2) rectangle (-2.7,-.3);
			\draw[black] (-2.7,-4.2) rectangle (-1.7,-.7);
			\draw[black] (-1.7,-4.2) rectangle (-0.7,-.4);
			\fill[black] (-0.7,-4.2) rectangle (0.3,2);
			\draw[pattern=north west lines, pattern color=black] (0.3, 0) rectangle (1.3,1);
			\draw[black] (0.3,-4.2) rectangle (1.3,1);
			\draw[black] (1.3,-4.2) rectangle (2.3,-0.5);
			\draw[black] (2.3,-4.2) rectangle (3.3,2.5);
			\draw[pattern=north west lines, pattern color=black] (2.3, 0) rectangle (3.3,2.5);
			\draw[black] (3.3,-4.2) rectangle (4.3,1);
			\draw[pattern=north west lines, pattern color=black] (3.3, 0) rectangle (4.3,1);

			\draw [decorate,decoration={brace,amplitude=3pt},xshift=-4pt,yshift=0pt]
			(-.7,0.02) -- (-.7,1.98) node [black,midway,xshift=-0.3cm] {$\Delta$};
		\end{scope}
	\end{tikzpicture}
	\end{subfigure}

	\caption{A visualization of the SVM loss for pointwise ranking}
	\label{fig:svm}
\end{figure}
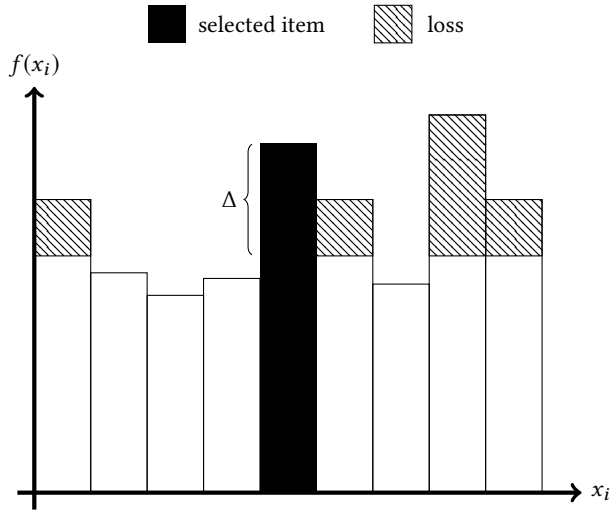

A visualization of this loss function is given in Figure~\ref{fig:svm}.
Each bar represents a possible suggestion, with the selected one being shown in black.
The y-axis displays how many points the model assigned to the respective suggestion.
The hatched areas show the SVM loss.
Everything above the selected suggestion as well as everything below it by a margin of $\Delta$ adds to the full loss.
Even though the selected suggestion had the second highest score, four suggestions contribute to the penalty in our example.

\subsection{Rprop}

Gradient descent is a natural choice for optimization in machine learning.
But since we never collected any data, we have no idea what gradient magnitudes for our problem are like.
Tuning the learning rate for vanilla gradient descent on the server prior to \fl thus does not work.

Adam~\cite{adam} and other optimization algorithms automatically adapt learning rates and are better suited for our problem setting.
We experimented with several such algorithms but finally settled for Rprop~\cite{Rprop}.
Rprop ignores gradient magnitudes and dynamically adapts learning rates for each weight individually.
The optimization algorithm bounds the magnitude of updates by design and thus limits by how much individual iterations affect the model.

Let $\eta_i^{(t)}$ be the step size for the $i$-th weight in the $t$-th iteration of gradient descent.
The value for the first and second iteration, $\eta_i^{(0)}$ and $\eta_i^{(1)}$, is a hyperparameter that needs to be chosen in advance.
This step size is then dynamically adapted for each weight, depending on the gradient.

The weights themselves are updated using
\begin{equation}
     \theta_i^{(t)} = \theta_i^{(t - 1)} - \eta_i^{(t - 1)} * \operatorname{sgn}\left(\frac{\partial E^{(t -
    1)}}{\partial \theta_i^{(t - 1)}}\right)
    \label{eq:Rprop}
\end{equation}
\noindent where the sign of the partial derivative of the error in the last step
with respect to the given weight is computed.
We go in the direction of descent using the determined step size.

In each iteration of Rprop, the gradients are computed and the step sizes are
updated for each dimension individually.
This is done by comparing the gradient's sign of the current and previous
iteration.
The idea here is the following:
\begin{itemize}
    \item When the signs are the same, we go in the same direction as in the
    previous iteration. Since this seems to be a good direction, the step size
    should be increased to go to the optimum more quickly.
    \item If the sign changed, the new update is moving in a different direction.
    This means that we just jumped over an optimum.
    The step size should be decreased to avoid jumping over the optimum again.
\end{itemize}

To implement this update scheme, the following formula is used:
\begin{equation}
    \eta_i^{(t)} = \begin{cases}
    \min(\eta_i^{(t - 1)} * \alpha, \eta_{\max}) & \text{if } \frac{\partial E^{(t)}}{\partial \theta_i^{(t)}} * \frac{\partial E^{(t - 1)}}{\partial \theta_i^{(t - 1)}} > 0 \\
    \max(\eta_i^{(t - 1)} * \beta, \eta_{\min}) & \text{if } \frac{\partial E^{(t)}}{\partial \theta_i^{(t)}} * \frac{\partial E^{(t - 1)}}{\partial \theta_i^{(t - 1)}} < 0 \\
    \eta_i^{(t - 1)} & \text{otherwise}
    \end{cases}
\label{eq:Rprop-lr}
\end{equation}
\noindent where $\alpha > 1 > \beta$ scale the step size, depending on whether
the speed should be increased or decreased. The step size is then clipped using
$\eta_{\min}$ and $\eta_{\max}$ to avoid it becoming too large or too small.
If a gradient was zero, a local optimum for this weight was found and the step
size is not changed.

There are well-known hyperparameters that tend to work well for Rprop~\cite{Rprop-empirical} and we validated those using simulations.
Rprop turned out to be a great choice for us for multiple reasons.
In contrast to Adam and most other gradient descent variants, it completely ignores the gradient magnitude.
It can thus deal with any data we might see after deploying our system.
Furthermore, it adapts learning rates dynamically for each weight individually, rendering the initial choice to not be that important.
A third reason is that the updates from Rprop are very interpretable and we can ensure that each iteration of optimization can only change frecency scores by a few points.

To get an additional compression advantage out of Rprop, it can be adapted to only use the signs of the gradient from each client.
Rather than computing the sign of the sum of gradients in Equation~\ref{eq:Rprop}, one can take the most common sign in the set of received gradients.
If we ignore the unlikely case of having a gradient of 0 in Equation~\ref{eq:Rprop-lr}, which would correspond to a perfect local optimum, clients now only need to share a single bit for each weight.
If one does not want to ignore this third case, two bits are required.
This is a strong compression factor, considering that we otherwise need to transfer 32 or 64 bits for each weight.
Additionally, this can also provide a privacy advantage because clients now need to share even less information with the server.

\subsection{Approximating Gradients}

Initially, we prototoyped our algorithms with computational graph libraries that can automatically compute gradients~\cite{tensorflow,pytorch-ad}.
However, it was difficult to add those to Firefox for our experiments.
The frecency algorithm is written in C++.
The corresponding modules are core to the browser and can only be changed if Firefox itself is updated.

To quickly prototype new ideas, Firefox has a separate mechanism for launching experiments, called \emph{Shield}~\cite{shield}.
This system lets us dynamically ship code to clients, completely independently of other major releases, by limiting which modules can be changed.
Since we wanted to use Shield for quick prototyping, we could not adapt the code behind the frecency algorithm itself.
However, it was still possible to change the weights behind the algorithm.

So instead of replacing the current implementation with a computational graph, we used a \emph{finite-difference method} for approximating the gradients.
If $g$ is any univariate function, its gradient can be approximated using:
\[
    g'(x) \approx \frac{g(x + \epsilon) - g(x)}{\epsilon}
\]
\noindent where $\epsilon > 0$ is a very small number.
To compute the gradient of a multivariate function, such as the SVM loss based on frecency, this process is then performed by iterating through all dimensions.
In each dimension, the value is changed by $\epsilon$ in the two directions, while all other values stay constant.
The resulting vector is our gradient estimate.

Because $\epsilon$ needs to be a small value for the approximation to be good, this formula inherently has problems with numerical stability.
To improve on this, we used an alternative that has previously been proposed for better stability~\cite{ad}:
\[
    g'(x) \approx \frac{g(x + \epsilon) - g(x - \epsilon)}{2 * \epsilon}
\]

It is worth noting that this numerical way of approximating gradients scales badly with the number of weights.
Instead of one forward pass in total to compute the gradient, we now need one forward pass per weight.
In our case, this was not a problem because there were comparatively few weights.
For neural networks with millions of parameters this would not be the case.

Still, this approach saved us a lot of engineering time, while only adding a very small performance penalty.
We did not have to rewrite any C++ code, and could treat the existing implementation as a black box.
It also makes our system generally applicable since anything in Firefox that has configurable weights can now be optimized.

\subsection{General Protocol and Server Side}

The server provides clients with a model file.
Clients fetch the current model when the browser is first opened as well as every time a new gradient descent iteration is completed.
This happens every 30 minutes and is triggered by the server.

Every time a user participating in the optimization performs a history or bookmark search in the URL bar, a gradient is computed.
This gradient is pushed to a Mozilla server using the Firefox Telemetry system~\cite{telemetry}, which has several advantages.
It is a well-designed system with clear rules about what can be collected.
There is a lot of infrastructure around using it and dealing with the data on the server.

All messages sent by clients are stored in a Parquet~\cite{parquet} data store.
A Spark streaming job~\cite{spark} reads the new updates from clients and averages them in real-time.
Every 30 minutes, the average update is then given to the optimizer, and applied to the model.
The resulting model is published and fetched by clients.

We store all gradient updates on the server for later analysis.
For a proven production system this is not strictly necessary.
The system could also work by adding gradients to the current average after which they are directly discarded.

\subsection{Safeguards}

The model that we are training is being used by the URL bar at the very same time as we are optimizing it.
Thus, it was important to make sure that the experiment does not degrade the quality of the URL bar too much if the optimization process fails.
To keep this from happening, we carefully configured Rprop and implemented several safeguards on top of it.

First of all, we initialize our model with the weights that the traditional frecency algorithm used.
This replicates the previous ranking behavior perfectly.
The optimization process thus starts off with a decent initial model.

To gradually improve on this initial solution, updates are bounded so that the model slowly converges to an optimum.
We wanted to avoid huge model changes to make it unlikely to jump far over an optimum.
Because our weights are intepretable, we can understand by how much updates can change frecency scores in one iteration, and limit this to a reasonable value.
In conjunction with the smart initialization, this ensures that our optimization process gradually improves the traditional frecency weights.

Since the weights of the frecency algorithm have a clear meaning, we were able to implement several additional constraints.
Recent visits should always be weighted higher than old ones.
To enforce this, the value of each time bucket weight is bounded to be smaller than the values of newer time buckets.
Finally, we force all weights to be nonnegative to ensure that visits can never be worth a negative number of points.

The exact safeguards we implemented for our system are highly domain-specific.
However, we expect that other domains should have similar constraints.
Bounding update size is a good idea, even if it is harder to interpret what exactly the weights represent.
Starting off from a decent initial model also helps to ensure that users do not interact with a bad model during the first few iterations of training.

\section{Experiment}
\label{sec:study}

\subsection{Simulations}
\label{subsec:simulations}

To prototype whether our ideas for optimizing the frecency weights could work, and to quickly iterate on them, a simulation was created before developing the actual system for Firefox~\cite{impl-simulations}.
This made it possible to simulate an entire \fl optimization process in little time, without having to wait for code to be deployed to actual clients.
Much of the code that we wrote for the simulation ended up being reused in Firefox~\cite{impl-client}.

The only major part that differs between the simulation and the actual implementation is what data is used for training.
Since no data should be collected, the simulation could not be based on real data from users.
Instead, a mock dataset was created.
This dataset was designed to resemble the data we expected users to generate.

Since there was no way of knowing how the data is actually distributed, several assumptions had to be made.
We modeled how recently websites were visited using a made-up distribution that was skewed towards more recent visits.
To decide on how many pages were bookmarked, we used existing statistics.

The frequency in which websites are visited is modeled using an exponential distribution with $\lambda = 7$.
This distribution mirrors the assumption that there are many websites that are only visited few times and few websites that are visited a lot.
For simplicity's sake, recency, type, and frequency were assumed to be independent of each other.

To model what suggestion a user clicks on, the existing frecency algorithm is used to compute a score for each suggestion.
Random noise, sampled from a normal distribution with $\mu = 0, \sigma^2 = 30$, is then added to the score.
The dataset assumes that the suggestion with the highest score is selected.
By using the existing frecency algorithm with some noise, it is easy to see whether the simulation finds useful weights, as they should be similar to the ones of the current algorithm.

It is worth noting that this dataset is likely to differ substantially from the data generated by real users.
It is difficult to perfectly describe how the data looks like without having seen any of it.
Still, creating the dataset allowed for quick prototyping, which made it much easier to make many design decisions.
We implemented variations of the optimization algorithms described in the previous section, and tested their properties using our simulations.
When the Firefox client-side and server-side implementations were ready, we used data from the simulation to test our final system end-to-end.

\subsection{Study Design}

After developing the Firefox client and server components of our system, 25\% of Firefox Beta users were enrolled in the experiment, which corresponds to roughly 500,000 daily active users.
Since it takes some time to roll out updates, only a part of the users was enrolled in the study before the optimization process was completed.

Users were partitioned into three groups:
\begin{enumerate}
	\item \emph{treatment}: The full study was shipped to these users. They compute updates, send them to the server, and start using a new model every 30 minutes.
	\item \emph{control}: This group is solely observational. No behavior in the URL bar actually changes. We are just collecting statistics for comparison to treatment.
	\item \emph{control-no-decay}: Firefox decays frecency scores over time. Our treatment group loses this effect because we are recomputing scores every 30 minutes. To check if the decay is actually useful, this group has no decay effect but uses the same original algorithm otherwise.
\end{enumerate}
60\% of users were assigned to the treatment group and 20\% to both control groups respectively.

To decide on these numbers, we performed a power analysis and used results from our simulation.
We carefully chose the number of people that should participate in the experiment for two reasons.
For one, if our study enrolls most Firefox users, we would block other studies that want to experiment with changes in the URL bar.
Another reason is that the experiment might break parts of Firefox.
If this happens, it should not affect unnecessarily many people.

Concretely, our analysis consisted of two parts:
\begin{enumerate}
    \item How many users do we need to have enough data to train a model? 
    (relevant for treatment)
    \item How many users do we need to show certain effects confidently? 
    (relevant for treatment and control)
\end{enumerate}

The first part was answered using simulations.
By using an adapted form of the simulation we used to decide on optimization hyperparameters, we could get some idea on how many users we would need.
Existing Telemetry data was helpful for this, as it allowed us to get some idea of how many history searches people perform every day~\cite{telemetry-selected-rank}.
The second part of the power analysis was tackled using classical hypothesis testing, based on the \mbox{\emph{Mann-Whitney-U test}}.
This analysis concluded that the control groups required fewer users than the treatment group.

To be able to evaluate how well our new model worked after the optimization converged, we also collected two metrics:
\begin{enumerate}
    \item The number of characters typed before selecting a result: Users should have to type few characters to find what they are looking for.
    \item The rank of the suggestion that was selected: The item that is selected should be as far on top as possible.
\end{enumerate}
Clients shared these two metrics with the server by sending them jointly with gradients.

\subsection{Analyzing the Results}

Over the course of the experiment, 723,581 users were enrolled in the study.
The model was fetched 58,399,063 times from the server.
360,518 users participated in sending updates and evaluation data to the server, accounting for a total of 5,748,814 messages.
The optimization phase of the experiment consisted of 137 iterations of 30 minutes each, or just under three days.
In this phase, 186,315 users sent pings to help in the training process.

A separate phase of purely evaluating the model was started afterwards and took a total of 10 days.
In this phase, 306,200 users sent 3,674,063 pings, which included statistics detailing how well the model worked for them.
Since all these users were assigned to treatment or control groups, the new model can be compared well to the old one that was used by the control groups.
Some users were enrolled but did not help with optimization or evaluation because they performed no history and bookmark searches.

\begin{figure}
    \centering
    \includegraphics[width=250px]{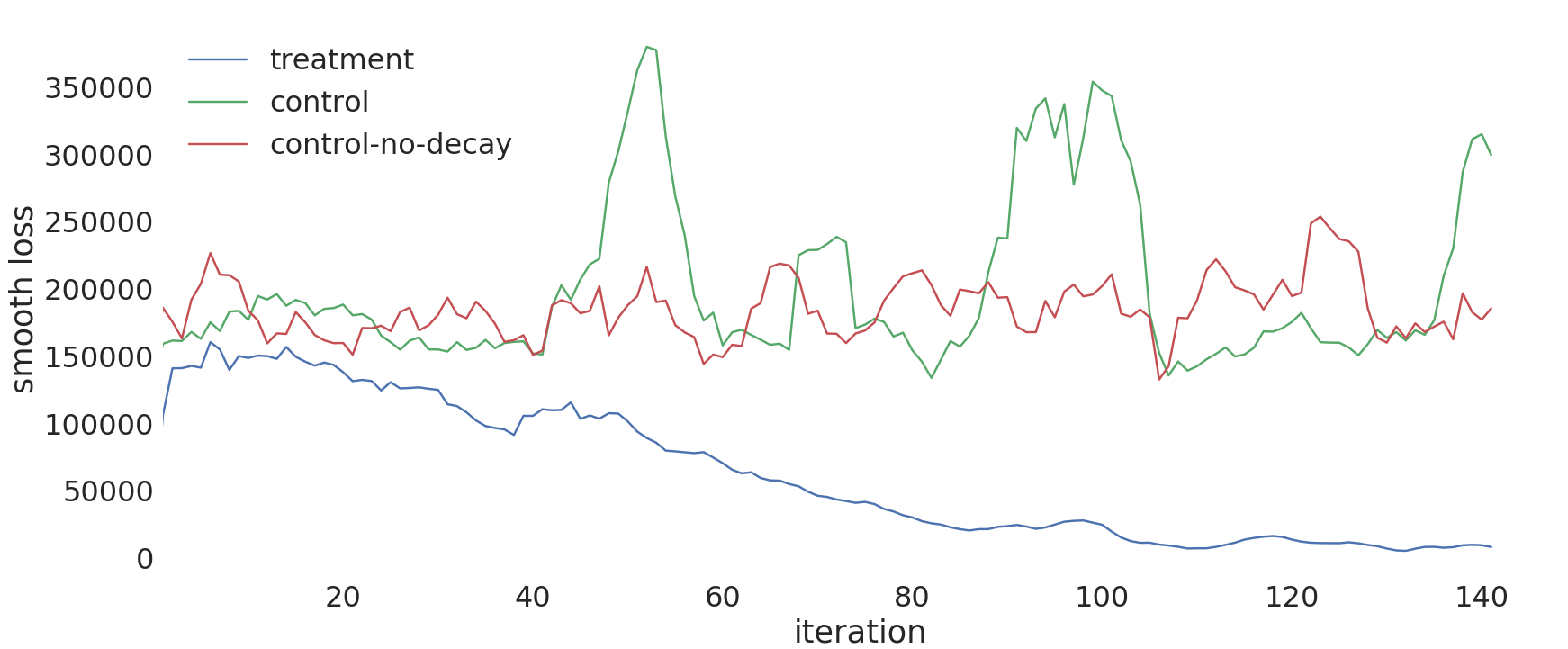}
    \caption{Rolling average of reported validation loss over the last 5 iterations}
    \label{fig:loss}
\end{figure}

During the optimization process, the loss of the model was supervised to check how well the training was going.
Figure~\ref{fig:loss} shows how the loss changed over time, across all three study variations.
There is some noise in this plot, since each iteration only had a very limited number of users.
However, it can still be seen that the loss of the treatment group continues to decline over the course of the experiment.
This shows that the optimization process generally worked.
After 40 iterations, less than one day of optimization, the loss of the treatment group is clearly below the loss of the control groups.

After the optimization process ended, an evaluation phase began to determine how well the new model works.
This is equivalent to the testing phase in machine learning.
The model is evaluated on new data that was not used for training or validation.
Table~\ref{fig:results} shows these results.
On average, users in the treatment group type about half a character less to find what they are looking for.
This is a strong improvement over both control groups.
However, users in the treatment group also choose suggestions that were ranked slightly worse.
Hypothesis testing determined that the changes in the treatment group were highly significant, with p-values being below $10^{-75}$.
Because we compared results of several experiment branches, we used a Bonferroni-corrected significance level of $\alpha = 0.05 / 6$.

\begin{figure}
	\centering
	\begin{tabular}{c||c|c}
		& \textbf{mean characters typed} & \textbf{mean rank chosen} \\
		\hline
		treatment & 3.6747 & 0.37435 \\
		control & 4.26239 & 0.35350 \\
		control-no-decay & 4.24125 & 0.35771
	\end{tabular}
	\caption{Results of the evaluation phase}
	\label{fig:results}
\end{figure}

From a user perspective, it is not clear if these changes improve the user experience.
While users now have to type a good amount less, they also select suggestions more often that are not on top of the list.
One potential explanation for this could be that the items they were looking for are displayed earlier in the suggestion list.
Since they spent less time typing, they might be willing to select an item that is not the top ranked one.
In the future, we plan to evaluate this hypothesis by collecting data about how the rank of the selected item changed while the user was typing.

It is difficult to determine purely based on the two metrics already collected if this change is good, since it is not clear how their importance should be weighted.
Instead, surveying users would be required to decide on which metric is more important.
But even if users are not satisfied with the new model, the Federated Learning system is still highly useful.
Since the optimization process works well, one would only need to find a loss function that correlates more closely with what users want.

\subsection{Analyzing the Optimization System}

\begin{figure}
    \centering
    \includegraphics[width=250px]{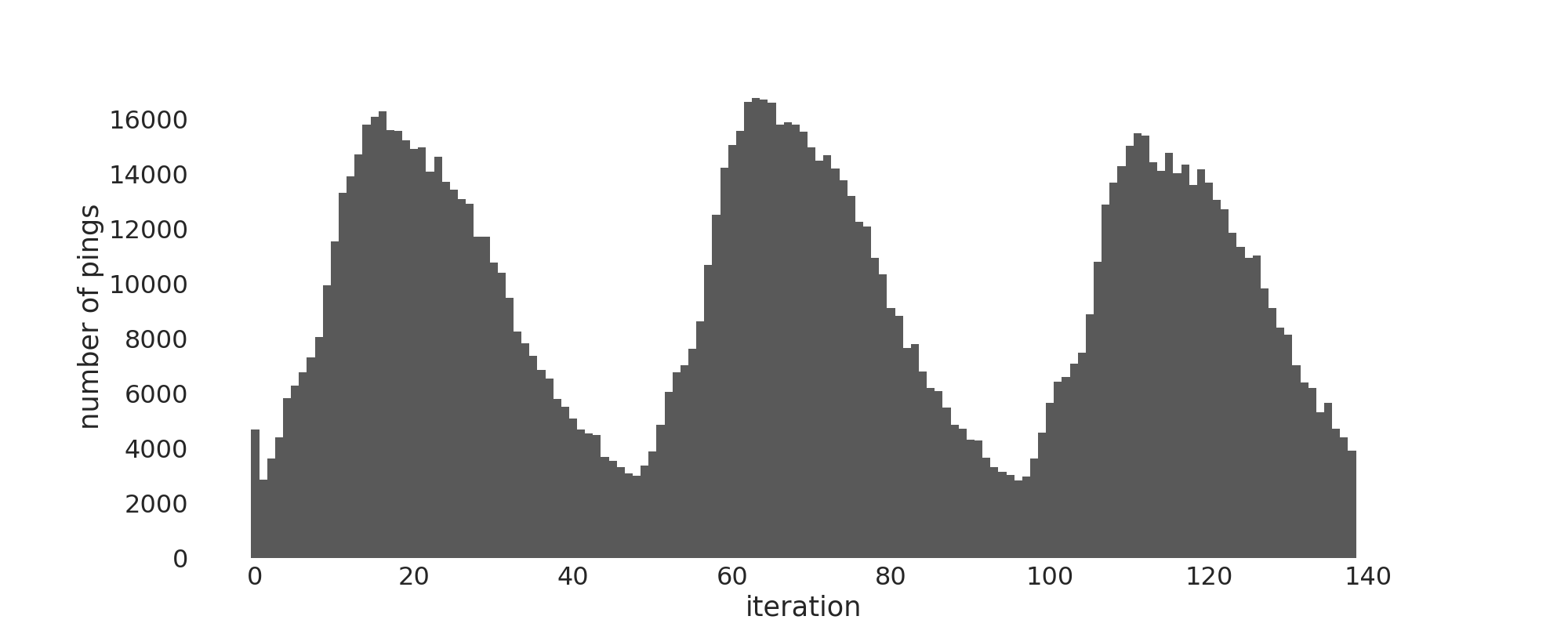}
    \caption{The number of pings sent by clients over time}
    \label{fig:pings}
\end{figure}

To learn from this experiment for further Federated Learning studies, we additionally analyzed all the update data later on.
In retrospect, the Federated Learning protocol we used was too simple.
Figure~\ref{fig:pings} shows how Firefox Beta activity in our study varies over time.
Since Firefox Beta usage has a bias towards Asian countries, we receive more pings during day time in Asia.

The protocol could be improved by dynamically determining the iteration length depending on how many updates were sent to the server so far.
This way, there would be no iterations with very few updates.
Furthermore, there could be more iterations during periods with many active users, allowing for a faster optimization process.

\begin{figure}
    \centering
    \includegraphics[width=250px]{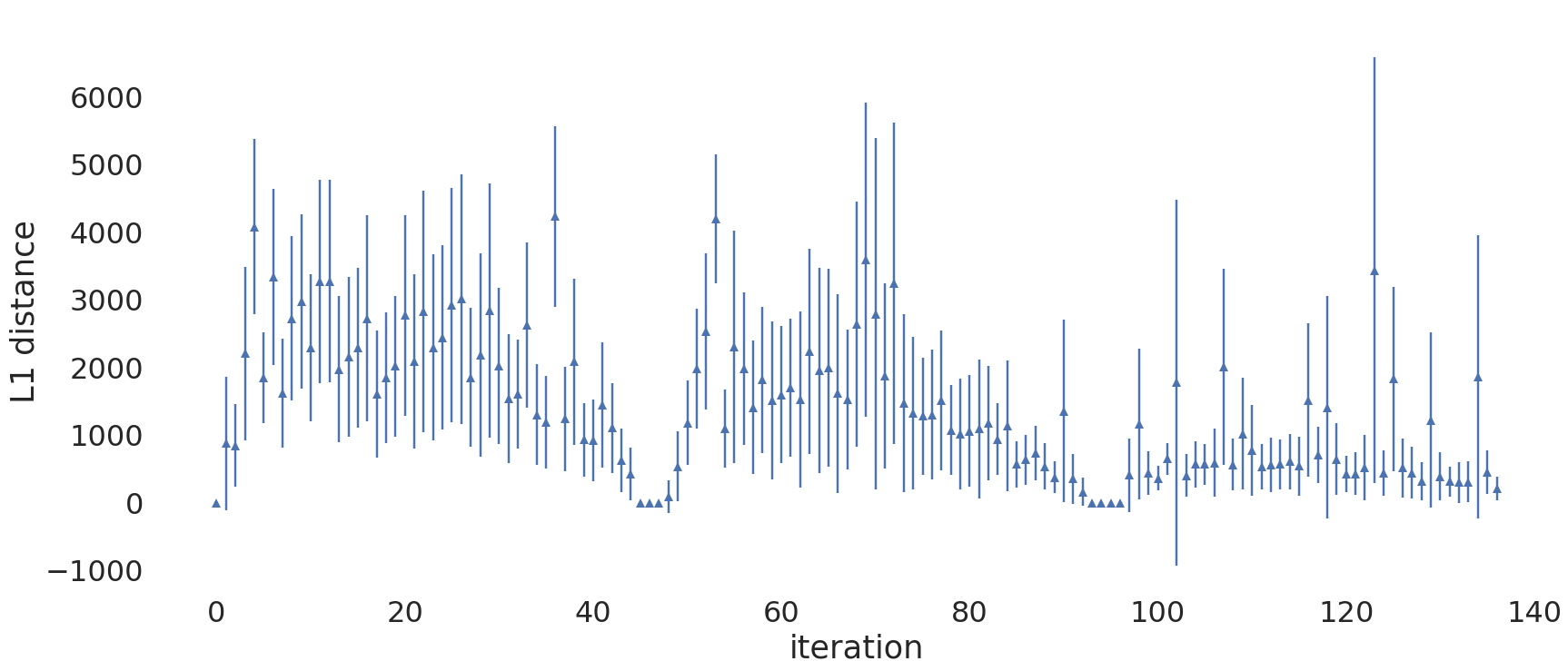}
    \caption{Mean and standard deviation of difference in update quality when using 2,000 updates}
    \label{fig:update-quality}
\end{figure}

A more sophisticated protocol could adapt the iteration length depending on how stable the current update estimate is.
We noticed that the later iterations of the optimization process require many fewer reports to compute a good estimate.
Figure~\ref{fig:update-quality} compares the update we actually used to updates we would get by randomly sampling 2,000 of these update reports.
The $L_1$-distance is used to perform this comparison.
Because of the randomness, the mean and standard deviation after 50 such simulations per iteration are reported.

It can be observed that the estimates become much more stable after iteration 100.
While the $L_1$-distance of two updates can be large without affecting the Rprop optimizer much, this is still an interesting result.
We observed similar results for the loss estimates.

The exact differences might be specific to our problem, but this observation can still be used to generally improve the system:
While updates are coming in, the server could check the variance of updates and start a new iteration earlier when it observes little variance.

\section{Conclusion}
\label{sec:conclusion}

In this paper, we introduced a \fl system built for use in Firefox.
Our system can optimize parts of Firefox purely based on user interactions.
The system is effective in optimization and preserves user privacy, as no personal data is ever shared with a server.
It is also widely applicable since it can replace or tune any heuristics, even if they can only be queried as black boxes.
We used the system to optimize the ranking of suggestion in the Firefox URL bar, which lead to users typing around half a character less before selecting an item.

Future work is multi-fold.
For one, the existing protocol can be improved to make better use of available data by adapting iteration length dynamically.
Furthermore, many other parts of Firefox can now be optimized using \fl.
The work here mostly lies in posing problems as learning tasks and in designing models.
Lastly, differentially-private mechanisms and secure multi-party aggregation could be added to our system in order to yield stronger privacy guarantees.


\begin{acks}



The authors would like to extend their thanks to several people at Mozilla:
Drew Willcoxon and Rob Helmer helped out with the Firefox client-side parts of the project.
Jeff Klukas and Katie Parlante provided additional support during the project.
Outside of Mozilla, we would like to thank Lukas Zilka and Ralf Hinze for giving valuable feedback while writing this paper.
\end{acks}

\bibliographystyle{ACM-Reference-Format}
\bibliography{one}

\end{document}